\documentclass[english,pdftex]{MetroSea2020}
\usepackage[T1]{fontenc}
\usepackage[latin1]{inputenc}
\usepackage{babel}[english]
\usepackage{hyperref}
\usepackage{flushend}
\usepackage[]{units}
\usepackage{graphicx}
\usepackage{authblk}

\usepackage[font={normalsize,it}]{caption}
\usepackage{fancyhdr}

\usepackage[capitalise]{cleveref}
\title{DANAE: a denoising autoencoder for underwater attitude estimation}
\author[1]{Paolo Russo}
\author[2]{Fabiana Di Ciaccio}
\author[2]{Salvatore Troisi}
\affil[1]{University of Rome La Sapienza, via Ariosto 25, Rome, paolo.russo@diag.uniroma1.it}
\affil[2]{Parthenope University of Naples, Centro Direzionale Isola C4, Naples, (fabiana.diciaccio, salvatore.troisi)@uniparthenope.it}

\begin{document}
\maketitle
\thispagestyle{imekopage}
\pagestyle{empty}
\begin{abstract}
One of the main issues for underwater
robots navigation is their accurate positioning,
which heavily depends on the orientation
estimation phase. The systems employed to this
scope are affected by different noise typologies,
mainly related to the sensors and to the irregular
noise of the underwater environment. Filtering
algorithms can reduce their effect if opportunely
configured, but this process usually requires
fine techniques and time. In this paper we propose DANAE, a deep Denoising AutoeNcoder for Attitude Estimation which works on Kalman filter IMU/AHRS data integration with the aim of reducing any kind of noise, independently of its nature.
This deep learning-based architecture showed to
be robust and reliable, significantly improving
the Kalman filter results. Further tests
could make this method suitable for real-time applications on navigation tasks. 
\end{abstract}

\section{Introduction} \label{sec:intro}
Localization is one of the most important tasks for unmanned robots, especially in underwater environments. Being a highly unstructured and GPS-denied environment, other than characterized by different noise sources and by the absence of man-made landmarks, the underwater setting provides more challenges for the orientation estimation. 

In a typical configuration, the Euler angles representing the vehicle attitude (roll, pitch and yaw) are obtained through the integration of raw data acquired by the sensors embedded into an Inertial Measurement Unit (IMU),
or in the more cost-effective Attitude and Heading Reference System (AHRS).

One of the most successful methods to perform this elaboration is based on the Kalman filter \cite{kalman1960} (KF) , in its linear and non linear versions \cite{st2004comparison}. 
Although known as the perfect estimator under some assumptions, the estimation provided by the KF strongly depends on a good knowledge of the error covariance matrices describing the noise affecting the system. Moreover, the on-line computation of these matrices is often required for anytime-varying or nonlinear system, squaring the number of necessary updating step at each time.
Finally, the procedure employed to accurately fine tune the filter parameters is known to be unintuitive,  requiring specific settings for different scenarios \cite{tereshkov2013intuitive}.

\begin{figure}
\centering
\includegraphics[scale=0.3]{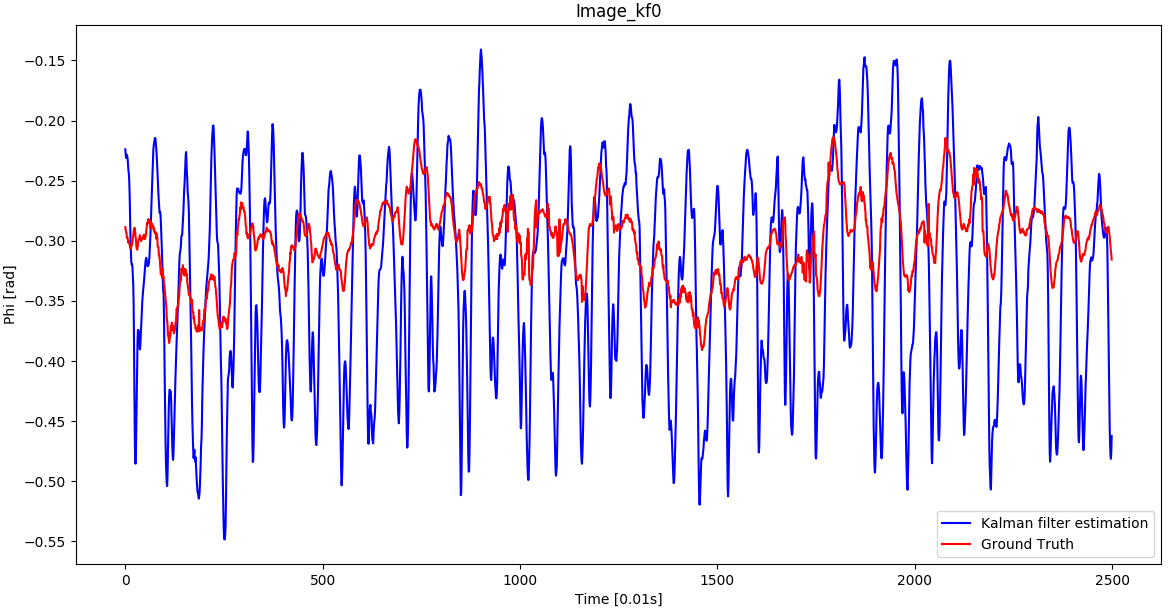}
\includegraphics[scale=0.3]{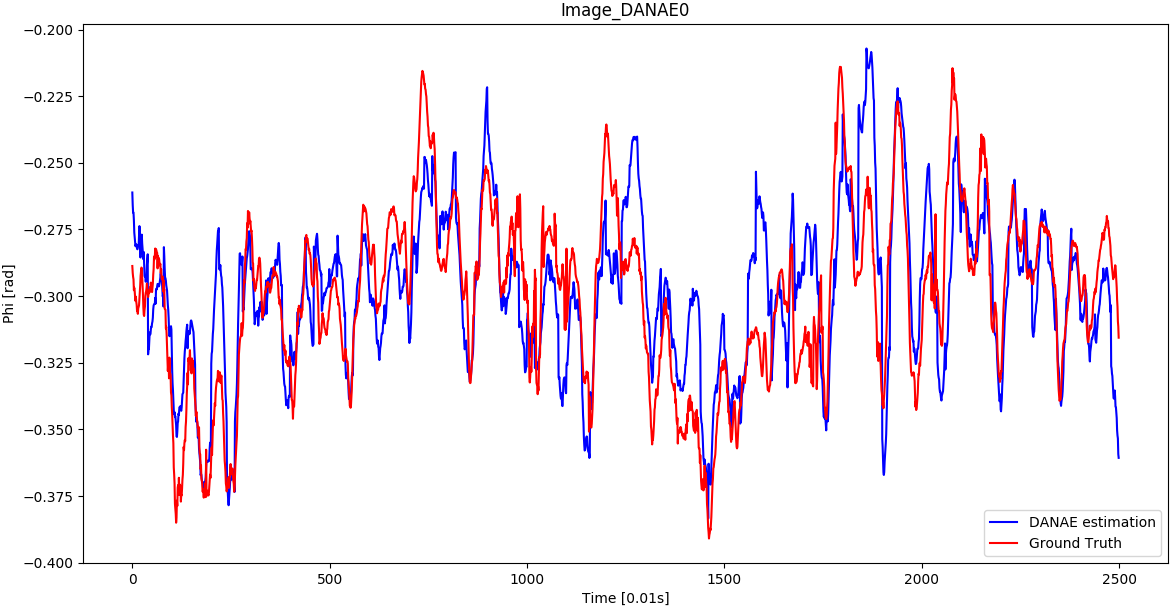}
\caption{KF (upside) and DANAE (downside) roll angle estimation compared to the ground truth. This experiment is made on a subsection of the slow walking set of OxIOD.}\label{fig:results}
\end{figure}

In order to overcome these issues, we propose a deep learning-based approach which aims to attenuate any source of error from the attitude estimation of a KF. The strength of our proposed method stands in the ability of acting as a full-noise compensation model for both noise and bias errors, without the need to separately process each influencing factor.   

Extensive tests performed on two different datasets to evaluate Euler angles show the power of our approach, with a sensible improvement of both mean squared error and max deviation w.r.t. ground truth data.



The remainder of this study is organized as
follows: \cref{sec:rel} presents a brief literature review for Kalman-based algorithms and deep learning techniques applied on similar tasks.
\cref{sec:method} introduces the concepts of KF and of deep denoising autoencoder, followed by a concise description of our method. In \cref{sec:data} the datasets used for the experiments are described. \cref{sec:setup} illustrates the details of our architecture and the hyper parameters chosen for the training. \cref{sec:results} analyzes the experiment results through the use of standard metrics, concluding with \cref{sec:conc} which contains some final considerations on the topic, including future possible improvements.

\section{Related works} \label{sec:rel}

The use of KF in robotic applications is ubiquitous. \cite{caron2006gps} developed a GPS/IMU multisensor fusion algorithm to increase the reliability of the position information, while \cite{li2013effective} and \cite{dicia2020kal} proposed an effective 
Kalman Filter which is able to exploit low-cost AHRS for efficient attitude estimation under various dynamic conditions, with an interesting underwater application developed by \cite{allotta2016new} and \cite{zhang2019application}.

Non-linear implementations of the filter include for example \cite{alatise2017pose}, which gets the robot pose by fusing camera and inertial data with an Extended Kalman Filter (EKF), and \cite{de2011uav} where the same task is accomplished using an Unscented Kalman Filter (UKF). For a detailed comparison between different kalman filters, see \cite{st2004comparison} or \cite{allotta2015comparison}.

Beside the critical task of sensor fusions, the estimation of both the sensors biases and noises is also crucial for an effective navigation system.
Nonetheless, Kalman filtering-based techniques constitute a powerful approach even to this problem. For example \cite{taghavi2016practical} exploited a kalman filter for accurate biases estimation in a distributed-tracking systems, while \cite{nirmal2016noise} identified and successfully removed the noise using a Kalman filter in a real time application. 



The rise of Deep Learning has radically changed fields like Computer Vision and Natural Language Processing. 
Since the spectacular success of \cite{krizhevsky2012imagenet},
Convolutional Neural Networks (CNN) produce state of the art accuracy on classification \cite{he2016deep}, detection \cite{he2017mask} and segmentation \cite{russo2019towards} tasks,
with Recurrent Neural Networks (RNN) being the backbone models for speech recognition \cite{graves2013speech} and sequence generation \cite{graves2013generating}.
Autoencoders are another successful deep architectures where the aim is to reconstruct a signal by learning a latent representation from a set of data. They have been used for realistic text \cite{baynazarov2019binary} and images \cite{sohn2015learning} generation; however, one of the most successful use of autoencoders is for noise removal.
Since their introduction \cite{vincent2008extracting}, denoising autoencoders (DAE) have been used for a broad number of tasks like medical images improvement \cite{gondara2016medical}, speech enhancement \cite{lu2013speech} and ECG signal boost \cite{xiong2016ecg}. 
To the best of our knowledge, DANAE is the first application of denoising autoencoder on attitude estimation.

\section{Method} \label{sec:method}
In this section we provide some basic concepts on linear KF and autoencoders followed by a brief overview of our method. However it should be emphasized that DANAE is filter agnostic and can be used seamlessly on non-linear KF (EKF, UKF) as well as any other type of filter able to perform attitude estimation.

\subsection{Kalman filter}
KF is a widely used algorithm for state estimation of dynamic systems since it is able to minimize the related variance under some perfect model assumptions.

The system behaviour in a discrete time setting can be described by a state equation \ref{eq:1} and a measurement equation \ref{eq:2}:
\begin{equation} \label{eq:1}
x_t=Ax_{t-1} + Bu_{t-1} + v_{t-1}
\end{equation}
\begin{equation} \label{eq:2}
y_t=Cx_t+ w_t 
\end{equation}

where $x_t$ is the state vector to be predicted, $x_{{t-1}}$ and $u_{{t-1}}$ are the state and the input vectors at the previous time step and $y_{t}$ represents the measurement vector. $A$ and $B$ are the system matrices and $C$ is the measurement matrix. The vectors $v_{t-1}$ and $w_t$ are respectively associated with the additive process noise and the measurement noise, assumed to be zero mean Gaussian processes.
The final estimate is obtained by a first prediction step (\ref{eq:3}, \ref{eq:4}) followed by the update phase (\ref{eq:5}, \ref{eq:6}, \ref{eq:7}, \ref{eq:8}):

\begin{equation} \label{eq:3}
x'_t=Ax_{t-1} + Bu_{t-1}
\end{equation}
\begin{equation} \label{eq:4}
P'_t=AP_{t-1}A^T+Q
\end{equation}
\begin{equation} \label{eq:5}
K_t=P'_t C^T (CP'_tC^T+R)^{-1}
\end{equation}
\begin{equation} \label{eq:6}
x_t=x'_t+K_t(y_t-Cx'_t)
\end{equation}
\begin{equation} \label{eq:7}
P_t=(I-K_tC) P'_t
\end{equation}

The a-posteriori state estimate $x_{t}$ is obtained as a linear combination of the a-priori estimate $x'_{t}$ and a weighted difference between the actual and the predicted measurements called residual (see equation \ref{eq:6}); the Kalman gain ($K$ in Eq. \ref{eq:5}) minimizes the a-posteriori error covariance ($P$ in Eq. \ref{eq:7}), initially set by the user. Finally, $Q$ and $R$ are the covariance matrices of the process and of the measurement noise. $Q$ models the dynamics uncertainty, while $R$ represents the sensors internal noises. 
These matrices heavily affect the final filter performance, thus a tricky tuning process is necessary to correctly estimate noises statistics. A proper fine-tuning is also important for sensors biases estimation; however, even in this case traditional approaches based on the KF suffer from implementation complexity and require non-intuitive tuning procedures \cite{tereshkov2015simple}.


\subsection{Denoising autoencoder}
DAE is a deep convolutional model which is able to recover clean, undistorted output starting from partially corrupted data as input. In the original implementation, the input data is intentionally corrupted thorough a stochastic mapping:
\begin{equation} \label{eq:8}
\tilde{x} \sim q_D(\tilde{x}|x)
\end{equation}

Then, the corrupted input is mapped into an hidden representation as in the case of a standard autoencoder:
\begin{equation} \label{eq:9}
h=f_\theta(\tilde{x})=s(W\tilde{x}+b)
\end{equation}
Finally, the hidden representation is mapped back to a reconstructed signal:
\begin{equation} \label{eq:10}
\hat{x}=g_{\theta'}(h)
\end{equation}
During the training procedure, the output signal is compared with a reference signal in order to minimize the L2 reconstruction error:
\begin{equation} \label{eq:11}
\mathcal{L}(x-\hat{x})=||x-\hat{x}||^2=||x-s(W\tilde{x}+b)||^2
\end{equation}
In our implementation, DANAE takes as input the noisy angles prediction performed by the KF and as reference signal the ground truth angles provided by the dataset.
For this reason, we underline that our method is able to remove both stochastic errors (e.g. electromagnetic- and thermo-mechanical- related ones), and systematic errors (due for example to sensors misalignment).






\section{Datasets} \label{sec:data}
DANAE has been developed and tested on the Oxford Inertial Odometry Dataset (OxIOD) \cite{chen2018oxiod} and on the Underwater Caves Sonar dataset (to which we will refer as UCS) \cite{mallios2017underwater}. 

OxIOD has been chosen for its accurate ground truth measurements over big heterogeneous settings. Developed for deep learning-based inertial-odometry navigation, OxIOD provides 158 sequences (for a total of 42.587 km) of inertial and magnetic field data acquired from low-cost sensors. Five users made indoor and outdoor acquisitions while normally walking with phone in hand, pocket and handbag and slowly walking, running and performing mixed motion modes. 
Different smartphones were used to acquire data, but the major part was collected by an iPhone 7 plus equipped with an InvenSense ICM20600. 
A Vicon motion capture system was used to get the ground truth 
with a precision down to 0.5 mm. 

The UCS dataset is collected by a Sparus AUV navigating in the underwater cave complex "Coves de Cala Viuda" in Spain. The vehicle explored two tunnels, closing a 500m-long path at a depth of approximately 20m. 
Among the equipped sensors (e.g. DVL, sonar, etc), a standard low cost Xsens MTi AHRS and an Analog Devices ADIS16480 are mounted.
The latter is a 10 DOF MEMS which provides more accurate raw sensors measurements and dynamic orientation outputs (obtained by their EKF fusion). 
The elaboration of images containing six traffic cones placed on the seabed allowed the relative positioning of the vehicle.
Unfortunately, the ground truth thus obtained is synchronized with the low-rate camera acquisitions, making the comparison with the high-rate IMU measurements inconsistent.
For this reason, we assumed that the orientation directly provided by the AHRS could at first glance substitute the true ground truth.
Despite not being a proper solution to the issue,
this choice allowed us to understand the ability
of DANAE to work in a true underwater scenario with its
unique features.

\section{Experimental setup} \label{sec:setup}
Some details on the experiments will be given in this section.
Both datasets have been split in a training and test set: in the case of OxIOD, for each setting we used the run $1$ as test set, leaving all the other sessions as training set. 
UCS provides instead a single file for each system containing all the measurements stored during the entire survey. We then decided to split the data, using the first 80\% to train DANAE and the remaining 20\% to test the performance.

Three main phases can be distinguished: during the first one, the inertial and magnetic field data are integrated with a linear KF, providing the estimation of the three Euler angles. In the second phase, these outputs are fed to the DANAE module for training, while on the third phase tests are performed using  a pipeline of KF and DANAE. All the code has been developed in Python 3.6.9 running on Ubuntu 18.04, with the help of Pytorch framework. 

\begin{table}
\caption{OxIO Dataset: performance of KF and DANAE vs GT. }
\label{table:resultsOxio}
\centering
\begin{tabular}{c | c c c}
\hline
KF  & $\phi$ & $\theta$ & $\psi$\\ [0.5ex]
\hline
Mean dev. [$rad$] & 0.0661 & 0.0483 & 1.9518 \\ [0.5ex]
Max dev. [$rad$]  & 0.2929 & 0.2134 &  9.0145 \\ [0.5ex]
RMSE  [$rad$]& 0.0815 & 0.0600 & 2.4000  \\ [1ex]
\hline
DANAE & $\phi$ & $\theta$ & $\psi$ \\ [0.5ex]
\hline
Mean dev. [$rad$] & 0.0224 & 0.0157 & 0.7392 \\ [0.5ex]
Max dev. [$rad$] & 0.1382 & 0.1082 &  5.7907 \\ [0.5ex]
RMSE [$rad$] & 0.0282 & 0.0196 & 1.3194 \\ [1ex]
\end{tabular}
\end{table}

\begin{table}
\caption{UCS Dataset: performance of KF and DANAE vs GT. Since the GT value of $\psi$ is not reliable, the corresponding results are not reported here.}

\label{table:resultsUCS}
\centering
\begin{tabular}{c | c c c}
\hline
KF  & $\phi$ & $\theta$ & $\psi$\\ [0.5ex]
\hline
Mean dev. [$rad$] & 0.0326 & 0.0328 & - \\ [0.5ex]
Max dev. [$rad$]  & 0.1476 & 0.1751 &  - \\ [0.5ex]
RMSE  [$rad$]& 0.0410 & 0.0412 & -  \\ [1ex]
\hline
DANAE & $\phi$ & $\theta$ & $\psi$ \\ [0.5ex]
\hline
Mean dev. [$rad$] & 0.0139 & 0.0147 & - \\ [0.5ex]
Max dev. [$rad$] & 0.0671 & 0.0769 &  - \\ [0.5ex]
RMSE [$rad$] & 0.0177 & 0.0190 & - \\ [1ex]
\end{tabular}
\end{table}

\subsection{KF initialization}
We implemented the filter in its most basic formulation following the equations from \ref{eq:3} to \ref{eq:7}. The covariance matrices P, Q and R have been initialized as identity matrix, and no tuning has been done with relation to both the internal system and the measurements noises. The elaboration of accelerometer and magnetometer raw data provided the measurements vector (see $Cx_t$ in \equationautorefname{2}), while the gyroscope-derived angles have been set as external input (see $Bu_{t-1}$ in \equationautorefname {1}).

\subsection{Autoencoder settings}
DANAE can work with any input signal length; here, we present the version working with a length of 20.  
The encoder part of DANAE is made of four \textit{dilated} 1D convolutions which bring the original 20-length signal to an hidden representation made of 128 features. The decoder part transforms this representation back to a 20-length vector by alternating three transposed-dilated 1D convolutions to four standard ones. While the transposed convolution is exploited to increase the input resolution back to the original size, the $i_{th}$ standard convolution is working on the sum of the $i_{th}$ encoder and the $i_{th}$ decoder outputs.  This approach, loosely inspired by the WaveNet architecture, is able to enforce additional constrains to the encoder/decoder pipeline, enabling a more faithful signal reconstruction.
All these layers have 128 $3x3$ kernels with an appropriate dilation value depending on the layer depth, while stride and padding have been fixed to 1. The Adam optimizer chosen for the training has been set with a fixed learning rate of 0.002 with a batch size of 16. The number of epochs has been set to 100 for UCS and to 1 for each set of OxIOD. Additional experiments performed with different hyper-parameters values did not produce any sensible difference in the final accuracy, demonstrating the robustness of our approach.

\section{Results} \label{sec:results}
\begin{figure}
\centering
\includegraphics[scale=0.3]{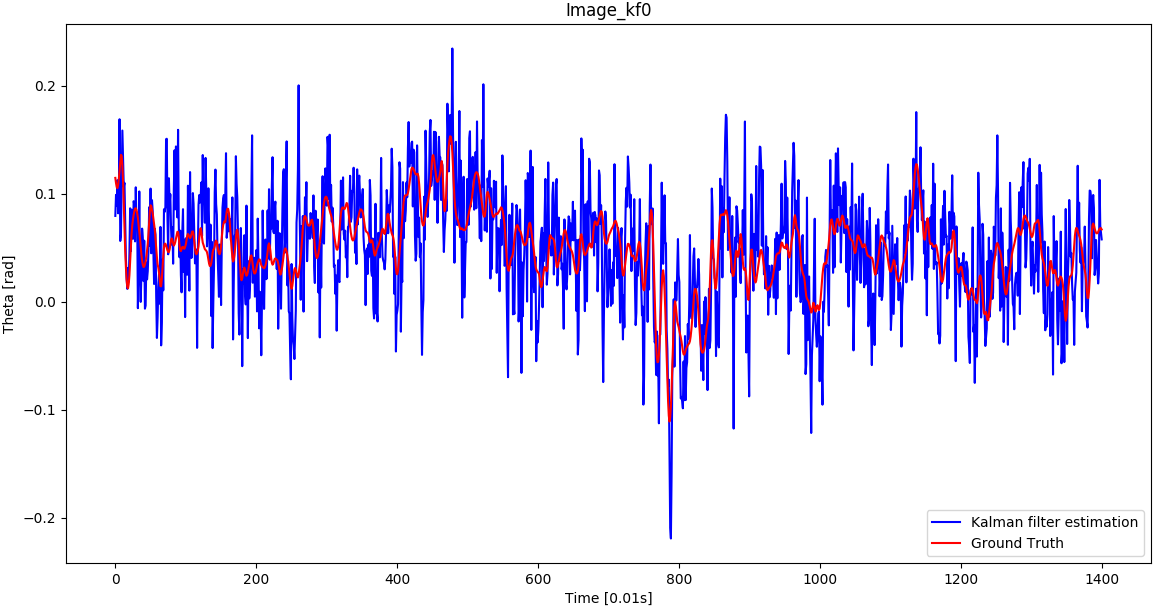}
\includegraphics[scale=0.3]{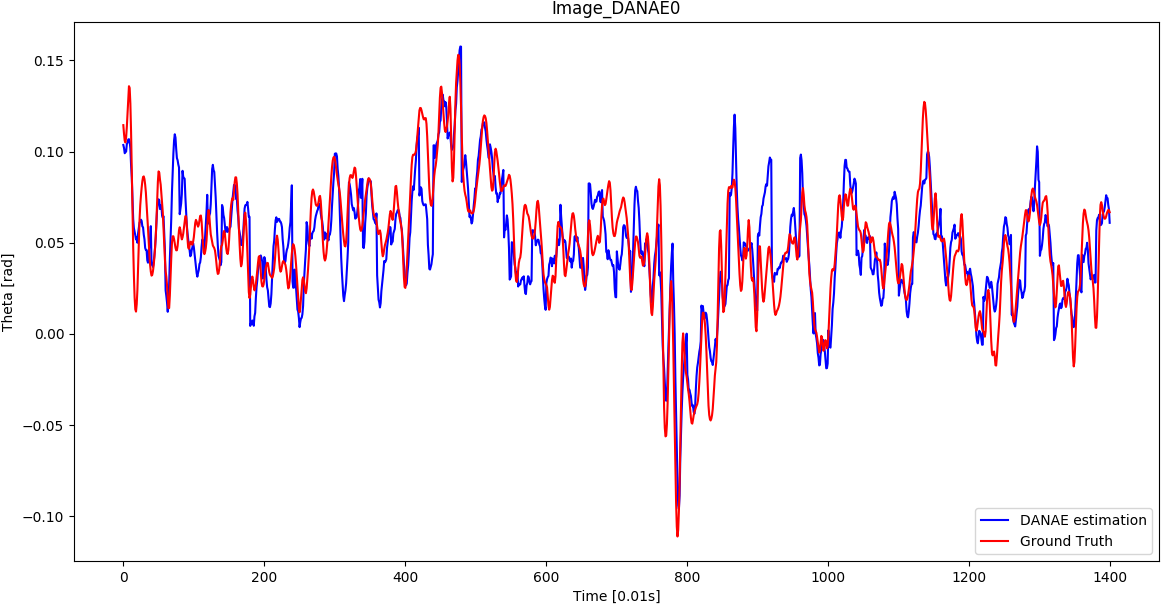}
\caption{KF (upside) and DANAE (downside) pitch angle estimation compared to the ground truth for the UCS dataset.}\label{fig:results_ucs}
\end{figure}
To numerically evaluate the DANAE performances, simple estimator as mean deviation, maximum deviation and root mean square error have been calculated with respect to the ground truth values and compared with those of the KF. 
As can be seen from Table \ref{table:resultsOxio} and Table \ref{table:resultsUCS}, DANAE is able to considerably improve the performance on all the estimators for both the datasets. Moreover, even though the strong noise affecting the KF predictions, DANAE exhibits a sensible lowering of the elongations for all the angles. In the case of OxIOD, DANAE provides a mean reduction of the RMSE equal to 63\%; Figure \ref{fig:results} shows the comparison between the KF and DANAE estimations for the $\phi$ angle w.r.t. the ground truth.

A similar result is also found in the UCS experiments: DANAE output faithfully resembles the reference signal for the estimated angles, with a final mean reduction of the RMSE equal to 55\%. Figure \ref{fig:results_ucs} reports the corresponding results for the $\theta$ angle.
Unfortunately, $\psi$ exhibits a perturbed behaviour both in the estimated and ground truth values, reason why numerical values are here omitted.
This can be probably related to erroneous sensors calibrations or to magnetometer effects, whose non-linearity results in a scale factor error. Moreover electromagnetic-produced deviations can considerably alter the estimations of this angle \cite{renaudin2010complete}.

\section{Conclusions} \label{sec:conc}
This paper proposes DANAE, the first denoising autoencoder model for attitude estimation. Our noise and filter agnostic architecture is able to compensate any kind of error with respect to a reference angle, boosting the KF prediction for both the considered datasets. Further experiments will be performed on non-linear filters (EKF and UKF), and deployments for on-line underwater applications will be investigated.

\bibliographystyle{ieeetr} 
\bibliography{biblio.bib}

\begin{thebibliography}{10}

\bibitem{kalman1960}
R.~E. Kalman, ``A new approach to linear filtering and prediction problems,''
  {\em Transactions of the ASME--Journal of Basic Engineering}, vol.~82,
  no.~Series D, pp.~35--45, 1960.

\bibitem{st2004comparison}
M.~St-Pierre and D.~Gingras, ``Comparison between the unscented kalman filter
  and the extended kalman filter for the position estimation module of an
  integrated navigation information system,'' in {\em IEEE Intelligent Vehicles
  Symposium, 2004}, pp.~831--835, IEEE, 2004.

\bibitem{tereshkov2013intuitive}
V.~M. Tereshkov, ``An intuitive approach to inertial sensor bias estimation,''
  {\em International Journal of Navigation and Observation}, vol.~2013, 2013.

\bibitem{caron2006gps}
F.~Caron, E.~Duflos, D.~Pomorski, and P.~Vanheeghe, ``Gps/imu data fusion using
  multisensor kalman filtering: introduction of contextual aspects,'' {\em
  Information fusion}, vol.~7, no.~2, pp.~221--230, 2006.

\bibitem{li2013effective}
W.~Li and J.~Wang, ``Effective adaptive kalman filter for
  mems-imu/magnetometers integrated attitude and heading reference systems,''
  {\em The Journal of Navigation}, vol.~66, no.~1, pp.~99--113, 2013.

\bibitem{dicia2020kal}
F.~Di~Ciaccio, S.~Gaglione, and S.~Troisi, ``A preliminary study on attitude
  measurement systems based on low cost sensors,'' {\em Proceedings of R3 in
  Geomatics Workshop (R3GEO), Communications in Computer and Information
  Science (CCIS), Springer}, in press.

\bibitem{allotta2016new}
B.~Allotta, A.~Caiti, R.~Costanzi, F.~Fanelli, D.~Fenucci, E.~Meli, and
  A.~Ridolfi, ``A new auv navigation system exploiting unscented kalman
  filter,'' {\em Ocean Engineering}, vol.~113, pp.~121--132, 2016.

\bibitem{zhang2019application}
X.~Zhang, X.~Mu, H.~Liu, B.~He, and T.~Yan, ``Application of modified ekf based
  on intelligent data fusion in auv navigation,'' in {\em 2019 IEEE Underwater
  Technology (UT)}, pp.~1--4, IEEE, 2019.

\bibitem{alatise2017pose}
M.~B. Alatise and G.~P. Hancke, ``Pose estimation of a mobile robot based on
  fusion of imu data and vision data using an extended kalman filter,'' {\em
  Sensors}, vol.~17, no.~10, p.~2164, 2017.

\bibitem{de2011uav}
H.~G. De~Marina, F.~J. Pereda, J.~M. Giron-Sierra, and F.~Espinosa, ``Uav
  attitude estimation using unscented kalman filter and triad,'' {\em IEEE
  Transactions on Industrial Electronics}, vol.~59, no.~11, pp.~4465--4474,
  2011.

\bibitem{allotta2015comparison}
B.~Allotta, L.~Chisci, R.~Costanzi, F.~Fanelli, C.~Fantacci, E.~Meli,
  A.~Ridolfi, A.~Caiti, F.~Di~Corato, and D.~Fenucci, ``A comparison between
  ekf-based and ukf-based navigation algorithms for auvs localization,'' in
  {\em OCEANS 2015-Genova}, pp.~1--5, IEEE, 2015.

\bibitem{taghavi2016practical}
E.~Taghavi, R.~Tharmarasa, T.~Kirubarajan, Y.~Bar-Shalom, and M.~Mcdonald, ``A
  practical bias estimation algorithm for multisensor-multitarget tracking,''
  {\em IEEE Transactions on Aerospace and Electronic Systems}, vol.~52, no.~1,
  pp.~2--19, 2016.

\bibitem{nirmal2016noise}
K.~Nirmal, A.~Sreejith, J.~Mathew, M.~Sarpotdar, A.~Suresh, A.~Prakash,
  M.~Safonova, and J.~Murthy, ``Noise modeling and analysis of an imu-based
  attitude sensor: improvement of performance by filtering and sensor fusion,''
  in {\em Advances in Optical and Mechanical Technologies for Telescopes and
  Instrumentation II}, vol.~9912, p.~99126W, International Society for Optics
  and Photonics, 2016.

\bibitem{krizhevsky2012imagenet}
A.~Krizhevsky, I.~Sutskever, and G.~E. Hinton, ``Imagenet classification with
  deep convolutional neural networks,'' in {\em Advances in neural information
  processing systems}, pp.~1097--1105, 2012.

\bibitem{he2016deep}
K.~He, X.~Zhang, S.~Ren, and J.~Sun, ``Deep residual learning for image
  recognition,'' in {\em Proceedings of the IEEE conference on computer vision
  and pattern recognition}, pp.~770--778, 2016.

\bibitem{he2017mask}
K.~He, G.~Gkioxari, P.~Doll{\'a}r, and R.~Girshick, ``Mask r-cnn,'' in {\em
  Proceedings of the IEEE international conference on computer vision},
  pp.~2961--2969, 2017.

\bibitem{russo2019towards}
P.~Russo, T.~Tommasi, and B.~Caputo, ``Towards multi-source adaptive semantic
  segmentation,'' in {\em International Conference on Image Analysis and
  Processing}, pp.~292--301, Springer, 2019.

\bibitem{graves2013speech}
A.~Graves, A.-r. Mohamed, and G.~Hinton, ``Speech recognition with deep
  recurrent neural networks,'' in {\em 2013 IEEE international conference on
  acoustics, speech and signal processing}, pp.~6645--6649, IEEE, 2013.

\bibitem{graves2013generating}
A.~Graves, ``Generating sequences with recurrent neural networks,'' {\em arXiv
  preprint arXiv:1308.0850}, 2013.

\bibitem{baynazarov2019binary}
R.~Baynazarov and I.~Piontkovskaya, ``Binary autoencoder for text modeling,''
  in {\em Conference on Artificial Intelligence and Natural Language},
  pp.~139--150, Springer, 2019.

\bibitem{sohn2015learning}
K.~Sohn, H.~Lee, and X.~Yan, ``Learning structured output representation using
  deep conditional generative models,'' in {\em Advances in neural information
  processing systems}, pp.~3483--3491, 2015.

\bibitem{vincent2008extracting}
P.~Vincent, H.~Larochelle, Y.~Bengio, and P.-A. Manzagol, ``Extracting and
  composing robust features with denoising autoencoders,'' in {\em Proceedings
  of the 25th international conference on Machine learning}, pp.~1096--1103,
  2008.

\bibitem{gondara2016medical}
L.~Gondara, ``Medical image denoising using convolutional denoising
  autoencoders,'' in {\em 2016 IEEE 16th International Conference on Data
  Mining Workshops (ICDMW)}, pp.~241--246, IEEE, 2016.

\bibitem{lu2013speech}
X.~Lu, Y.~Tsao, S.~Matsuda, and C.~Hori, ``Speech enhancement based on deep
  denoising autoencoder.,'' in {\em Interspeech}, pp.~436--440, 2013.

\bibitem{xiong2016ecg}
P.~Xiong, H.~Wang, M.~Liu, S.~Zhou, Z.~Hou, and X.~Liu, ``Ecg signal
  enhancement based on improved denoising auto-encoder,'' {\em Engineering
  Applications of Artificial Intelligence}, vol.~52, pp.~194--202, 2016.

\bibitem{tereshkov2015simple}
V.~M. Tereshkov, ``A simple observer for gyro and accelerometer biases in land
  navigation systems,'' {\em The Journal of Navigation}, vol.~68, no.~4,
  pp.~635--645, 2015.

\bibitem{chen2018oxiod}
C.~Chen, P.~Zhao, C.~X. Lu, W.~Wang, A.~Markham, and N.~Trigoni, ``Oxiod: The
  dataset for deep inertial odometry,'' {\em arXiv preprint arXiv:1809.07491},
  2018.

\bibitem{mallios2017underwater}
A.~Mallios, E.~Vidal, R.~Campos, and M.~Carreras, ``Underwater caves sonar data
  set,'' {\em The International Journal of Robotics Research}, vol.~36, no.~12,
  pp.~1247--1251, 2017.

\bibitem{renaudin2010complete}
V.~Renaudin, M.~H. Afzal, and G.~Lachapelle, ``Complete triaxis magnetometer
  calibration in the magnetic domain,'' {\em Journal of sensors}, vol.~2010,
  2010.

\end{thebibliography}


\end{document}